\documentclass[runningheads]{llncs}
\usepackage[T1]{fontenc}
\usepackage{siunitx}
\usepackage{amsmath}
\usepackage{comment}
\usepackage{graphicx}
\usepackage{url}

\begin{document}
%
%
\title{A Deep Dive into Axiomatic Design - Part I: Problem Formulation}
\titlerunning{}  
%
\author{Aydin Homay}
\authorrunning{Aydin Homay} 
%
%
\institute{Technische Universität Dresden, Dresden 01069, Germany,\\
\email{homay@ieee.org}\\
\texttt{https://orcid.org/0000-0002-6425-7468}
}

\maketitle              

\begin{abstract}
Problem formulation---translating customer needs and constraints into a minimum set of independent first-level functional requirements—is arguably the most critical step in every design framework, including axiomatic design; yet it is frequently misunderstood or underestimated in practice. This paper focuses exclusively on problem formulation in axiomatic design: it clarifies what first-level FRs are (and are not), explains why they should not legitimately vary across designers given the same needs and constraints, and highlights intrinsic difficulties and recurring pitfalls that lead to design failure. The discussion is grounded primarily in Nam P.~Suh's three books---\emph{The Principles of Design}, \emph{Axiomatic Design: Advances and Applications}, and \emph{Complexity Theory}---and it offers practical guidance to help designers formulate well-posed first-level FRs. Finally, the paper briefly revisits problem formulation in the era of large language models and discusses what such tools can (and cannot) contribute at the first level.
\keywords{axiomatic design, abstract thinking, problem formulation.}
\end{abstract}
\section{Introduction}
As a crisper version of John von Neumann’s well-known quote, cited by Jackson~\cite{Jackson-1995}: ``There is no sense in being precise when you don’t know what you are talking about.''

Regardless of whether the statement above is an accurate interpretation of what von Neumann~\cite{Von-Neumann-1944} meant by poorly formulated economic problems, the underlying lesson still applies: being overly detailed at the start of the design process—i.e., during problem formulation in terms of requirements—obscures uncertainty, freezes premature design decisions, and creates an illusion of understanding what needs to be done. Ultimately, this increases the risk of misalignment and costly changes later in development.

Axiomatic design (AD) provides a unique process in which requirements formulation is central, and maintaining problem formulation in terms of requirements (i.e., functional requirements) in a solution-neutral environment is a must. But this can be achieved once the process starts from highly abstract (i.e., high-level) requirements and proceeds through decomposition via an iterative interplay between the problem space and the solution space (i.e., the ``what'' and the ``how''). Such a process allows sub-requirements to emerge in response to progressively clarified design decisions~\cite{Suh-1990,Jackson-1995}. However, the literature on AD presents recurring misconceptions regarding this step.

For example, in~\cite[pp.~167]{Liu-2021}, Ang Liu claims that, depending on the experience, background, and knowledge of the designer, a variety of different problems can be formulated for the same given set of customer needs and design constraints.

He supports his argument by providing the following example: ``in order to treat cancer, one may define it as a biological problem that should be solved by designing a biological experiment to understand the genesis and propagation of cancer. Alternatively, another person may define the problem as a pharmaceutical problem of designing a new drug to retard cancer development. Another individual may characterize it as a public healthcare problem and formulate it as a public campaign to discourage smoking.'' 

However, as discussed below, this interpretation is misleading because it does not apply to the first-level FRs, which are obtained by directly translating customer needs and constraints (specifically, input constraints). Rather, it is more appropriate for lower-level FRs that can vary depending on the design parameters (DPs) chosen by different designer’s with different backgrounds and expertise~\cite[pp.~56]{Suh-Cavique-Timothy-2021};~\cite[pp.~61]{Suh-Cavique-Timothy-2021}. 

Note that in Liu’s example, one can argue that ``treat cancer'' is, in fact, a first-level FR arising from problem formulation for a business need, such as lowering mortality rates as a societal objective within the business domain, whereas the \emph{biological}, \emph{pharmaceutical}, or \emph{public health} framings correspond to alternative design parameters (i.e., alternative solution strategies) to enable ``cancer treatment'' as part of customer needs satisfaction. 

It is important to note that while Suh argues that multiple solutions may exist for a given set of requirements, all of which can satisfy those requirements, this study found no statement across his three books~\cite{Suh-1990,Suh-2001,Suh-2005} claiming that, for the same customer needs and constraints, first-level FRs may legitimately vary from one designer to another. In fact, such a claim would conflict with the notion of system definition: every system ultimately has a set of essential elements, properties, and structures that determine its function and identity~\cite[pp.~5]{Bossel-2007}, necessitating a common understanding of customer needs. This is especially important in the construction of systems that require cross-functional teams to form a shared understanding of customer needs, regardless of designers' backgrounds or prior knowledge. 

For example, consider a \textit{radiation therapy} system for cancer treatment that is developed based on customer needs by engineers and scientists from different disciplines (e.g., optical physics, software engineering, electrical engineering, mechanical engineering, and medical physics). Despite their different backgrounds and expertise, they must share a common understanding of ``what'' the system must accomplish (i.e., a common understanding of the problem) if they want to successfully develop the product. 

Therefore, problem formulation must be expressed at a level of abstraction that enables practitioners from different backgrounds to make consistent judgments about what is required, yielding a shared set of essential requirements (i.e., a common problem formulation) on which product development can reliably proceed.

Throughout this study, functional requirements formulated at this level are referred to as \emph{essential FRs} because this term better conveys their foundational role than the label ``first-level FRs.'' All other FRs are referred to as \emph{derived FRs}, since their existence depends on the chosen ``how'' by which the essential FRs are satisfied.

The rest of this paper is structured into five main parts. Section~\ref{background} provides a brief introduction to axiomatic design theory. Section~\ref{intransic_difficulty} examines the intrinsic difficulties of problem formulation and summarizes recurring pitfalls. Section~\ref{abstract_thinking} discusses the role of abstract thinking during essential-FR formulation. Section~\ref{advancement} revisits problem formulation in the era of large language models and outlines what such tools can (and cannot) contribute at the first level. Finally, Section~\ref{Conclusion} concludes the paper.

\section{Background}
\label{background}
\begin{figure}[h]
    \centering
    \includegraphics[width=0.7\textwidth]{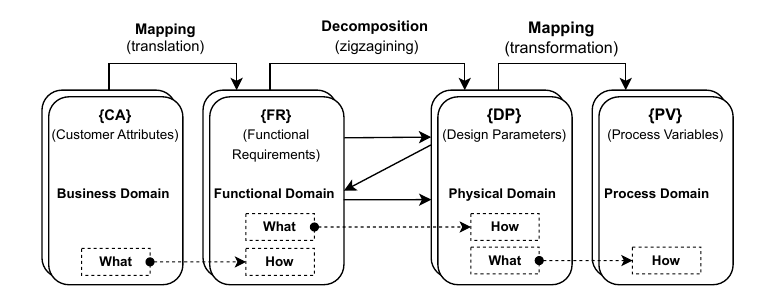}
    \caption{Mapping between axiomatic design domains (``what'' to ``how'').~\cite[pp.~11]{Suh-2001}}
    \label{fig:axiomatic_design_domains}
\end{figure}

Axiomatic design (AD) comprises four primary domains (see Fig.~\ref{fig:axiomatic_design_domains}). The design process starts from the \emph{customer domain}, where customer needs are captured as customer attributes (CAs). These CAs are then translated, in a solution-neutral environment, into a minimal set of independent requirements named functional requirements (FRs) (hereafter, essential FRs) in the \emph{functional domain}, subject to input constraints~\cite[pp.~22]{Suh-2005}. Next, the \emph{physical domain} defines the design parameters (DPs) that satisfy the FRs. Finally, the \emph{process domain} specifies the process variables (PVs) required to generate the selected DPs.

Throughout the design process, the designer must satisfy two design axioms. The \emph{Independence Axiom} ensures that the design is either \emph{uncoupled} or \emph{decoupled} , and the \emph{Information Axiom} chooses the best design (i.e., the design with the highest probability of success) among candidate designs (i.e., sets of DPs).

\subsection{Constraint}
Constraints (CS) are the limiting values, conditions, or bounds on acceptable solutions. Constraints are categorized as \textit{input constraints} or \textit{system constraints}. Input constraints, specific to the design goals, are imposed through design specifications and must be satisfied by all proposed designs~\cite[pp.25]{Suh-2005}. System constraints arise from the operational environment of the design and reflect decisions made during the design process in the physical domain~\cite[pp.25]{Suh-2005}.

\subsection{Zigzagging}
\label{zigzagging}
Zigzagging is an iterative decomposition technique in which the designer alternates between the ``what'' domain (e.g., the functional domain) and the ``how'' domain (e.g., the physical domain). For example, starting with the essential \( FR_1 \), the associated \( DP_1 \) is identified, followed by the refinement of the subrequirements (\( FR_{1.1} \), \( FR_{1.2} \)) based on the chosen \( DP_1 \). This process continues until no further breakdown is needed, forming a hierarchical tree structure~\cite[pp.29-30]{Suh-2001}. At each level, the design equation (e.g., product design equation ${FR} = [A]{DP}$ when interplaying between the functional domain and the physical domain) must be written down to ensure the Independence Axiom is satisfied~\cite[pp.29]{Suh-2001}.

\section{Intrinsic Difficulty of Problem Formulation}
\label{intransic_difficulty}
Problem formulation starts by translating customer needs into a minimum set of independent \textbf{essential} FRs. This step is the most critical and challenging step in axiomatic design~\cite[pp.~30]{Suh-1990}~\cite[pp.~10]{Suh-2001} and is completed when the \emph{design range} for the FRs, along with the relevant constraints, has been defined~\cite[pp.~15]{Suh-2001}~\cite[pp.~96]{Suh-2001}.

When the problem is formulated poorly, the resulting design is likely to be unacceptable or unnecessarily complex~\cite[pp.~30]{Suh-1990};~\cite[pp.~203]{Suh-2021-FR}. Yet, in practice, this step is often overlooked, misunderstood ~\cite{Liu-2021}, or executed inadequately (for example, formulating FRs from user requirements of the system rather than from business needs~\cite{Pourabbas-Pecoraro-2021}).

This study argues that, although AD is a systematic design framework with axioms, corollaries, and theorems to guide designers throughout the entire design process, the problem formulation step (i.e., transitioning from the business domain to the functional domain) remains a significant challenge and is easy to get wrong. 

The main reason is that the \emph{Independence Axiom}~\cite[pp.~16]{Suh-2001} and the \emph{Information Axiom}~\cite[pp.~39]{Suh-2001}, which are intended to be the two most powerful tools to guide the design process, can only be applied once the essential FRs are established; only then can the designer use them to evaluate the design. The same applies to design equations (product design~\cite[pp.~18]{Suh-2001} and process design~\cite[pp.~19]{Suh-2001}). None of these equations are really meant to guide the designer in translating business needs into the essential FRs.

Put differently, although axioms and design equations offer ``deductive'' means to evaluate the design layer-by-layer, translating customer needs into essential FRs arguably depends on ``inductive''\footnote{Non-necessary inferences, which constitute a somewhat heterogeneous class that relies solely on statistical information, such as observed frequencies of a given trait within a fully observable population~\cite{Douven-2025}.} and ``abductive''\footnote{Non-necessary inferences, which typically begin with an incomplete set of observations and proceed to the likeliest possible explanation for the set~\cite{Douven-2025}.} inferences.

In ``deductive'' inferences, what is inferred is  necessarily true if the premises from which it is inferred are true; that is, the truth of the premises guaranties the truth of the conclusion~\cite{Douven-2025}. For example, if the design matrix is diagonal, the design is \emph{uncoupled}; if it is triangular, the design is \emph{decoupled}; otherwise, the design is \emph{coupled} (see Eq.~\ref{design_matrix}).

\begin{equation*}
\setlength{\arraycolsep}{3pt}
\renewcommand{\arraystretch}{1.0}
\begin{array}{c@{\qquad}c@{\qquad}c}
\displaystyle [A]=\begin{bmatrix}
\times & 0 & 0\\
0 & \times & 0\\
0 & 0 & \times
\end{bmatrix}
&
\displaystyle [A]=\begin{bmatrix}
\times & \times & \times\\
0 & \times & \times\\
0 & 0 & \times
\end{bmatrix}
&
\displaystyle [A]=\begin{bmatrix}
\times & \times & 0\\
\times & \times & \times\\
0 & \times & \times
\end{bmatrix}
\\[1.5em]
\text{Uncoupled (diagonal)}
&
\text{Decoupled (triangular)}
&
\text{Coupled (otherwise)}
\end{array}
\label{design_matrix}
\end{equation*}

By contrast, problem formulation does not enjoy such comfort and provides far less certainty. Designers frequently have to state the problem using a combination of complete and incomplete observations in the customer/business domain. At this point, the design process involves not only synthesis but also a combination of synthesis and analysis. 

Once design has progressed from the functional domain to the physical domain, the task becomes more of a ``deductive'' matter, where design can be continued bidirectionally. 

\subsection{Bidirectional Design Process}
In a bidirectional design process, designers typically carry out \emph{analysis} and \emph{synthesis} together~\cite{Takeda-1992}. If the mapping is straightforward, i.e., there is a linear causality between the ``what'' and the ``how,'' then the \emph{design knowledge}\footnote{The union of knowledge in the ``what'' domain with knowledge in the ``how'' domain—forms the design knowledge—} is already present in the form of a clear understanding of the ``what needs to be achieved'' with a known best solution as the ``how the what will be achieved'' (e.g., if a vehicle is intended to operate on British motorways, problem formulation must result in an FR that specifies a right-hand steering configuration); in that case, the design is mainly synthesis and can proceed deductively. Otherwise, designers must jointly analyze the ``what'' and synthesize a plausible ``how''; i.e., the designer should discover the knowledge of ``what'' and the knowledge of ``how'' incrementally.

\subsection{Different Types of Knowledge}
Therefore, this study maintains that the design knowledge involved in such a mapping process (e.g., from a non-linear “what” to the “how”) can be formulated in general forms as

\[
K_W \cup K_H \vdash K_D
\]

where $K_W$ represents the knowledge of the ``what'' that is desired (e.g., information representing customer preferences and constraints ~\cite[pp.~51]{Suh-2001}), $K_H$ represents the knowledge of the ``how'' (e.g., potential essential FRs or DPs in the case of interplay between the functional domain and the physical domain), and $K_D$ denotes the total available knowledge (i.e., design knowledge). 

In contrast to essential FRs, derived FRs\footnote{FRs that do not belong to the set of essential FRs.} are obtained by “zagging” from the physical domain back to the functional domain and therefore intrinsically have access to the complete design knowledge at the particular level of decomposition (i.e., $K{_D{_x}}$). The translation of customer needs into essential FRs (i.e., problem formulation) is, and must be, initiated using only the initial $K_W$ in the business domain.

The initial $K_W$ (i.e., the ``what'' knowledge in the business domain) is obtained by processing information in the business domain in the form of customer attributes (i.e., needs) and input constraints, relying on either ``inductive'' or ``abductive'' reasoning. The result is supposed to be \emph{a minimum set of independent FRs}, which the study describes as essential-FRs. 

To develop the corresponding ``how'' knowledge (i.e., $K_H$), the designer must perform ``zigging'' into the physical domain to conceptualize solutions (e.g., design parameters) and form the $K_H$ by choosing appropriate DPs.

Note that during the ``zagging'' from the ``how'' domain back to the ``what'' domain, the available ``how'' knowledge $K_H$ may increase as a result of capturing emergent system constraints ($C_{\text{sys}}$). This results in a reduction of scope in the subsequent ``what'' domain and more detailed ``what'' in lower levels of the decomposition tree (i.e., the $K_W{_{n+1}}$ will be restricted). In general form

\[
K_W' = K_W \cap C_{sys},
\]
and the new total available design knowledge (i.e., $K_D$) can be expressed as
\[
K_H \cup \bigl( K_W \cap C_{sys} \bigr) \;\vdash\; K_D.
\]

From this stage on, $K_D$ is updated step by step as the design is decomposed. This gradually constrains both the “what” and the “how”—that is, it restricts the available design decisions—which, in turn, helps curb arbitrariness (i.e., randomness) in the overall design process and mitigates the so‑called seductive flexibility in software systems~\cite{Booch-2007}. It generally increases certainty in the design decisions being made as the design advances step by step.

\section{Essential Functional Requirements}
Essential FRs emerge primarily from $K_W$ through problem formulation; they define the boundary of the design in the functional domain and anchor the independence axiom at the highest abstraction level. In contrast, all subsequent i.e., derived FRs are generated under the influence of $K_D$ because they arise from iterative zigzagging, where the change in high-level DPs impacts the derived FRs but not the essential FRs. Meanwhile, the change in the essential FRs will impact the chosen essential DPs and consequently lead to a rippling change in all design decompositions.

This ``ontological asymmetry in design decomposition'' arises only between essential FRs and first-level DPs: changes to the essential FRs affect the selected set of DPs, but not vise versa.

Accordingly, derived FRs can be viewed as largely ``solution-contingent'' and thus epistemological, whereas essential FRs are comparatively ``solution-neutral'' and therefore ontological. The following subsection will examine this in greater detail.

\subsection{What Makes Essential FRs Distinct}
\begin{figure}[ht]
    \centering
    \includegraphics[width=0.8\linewidth]{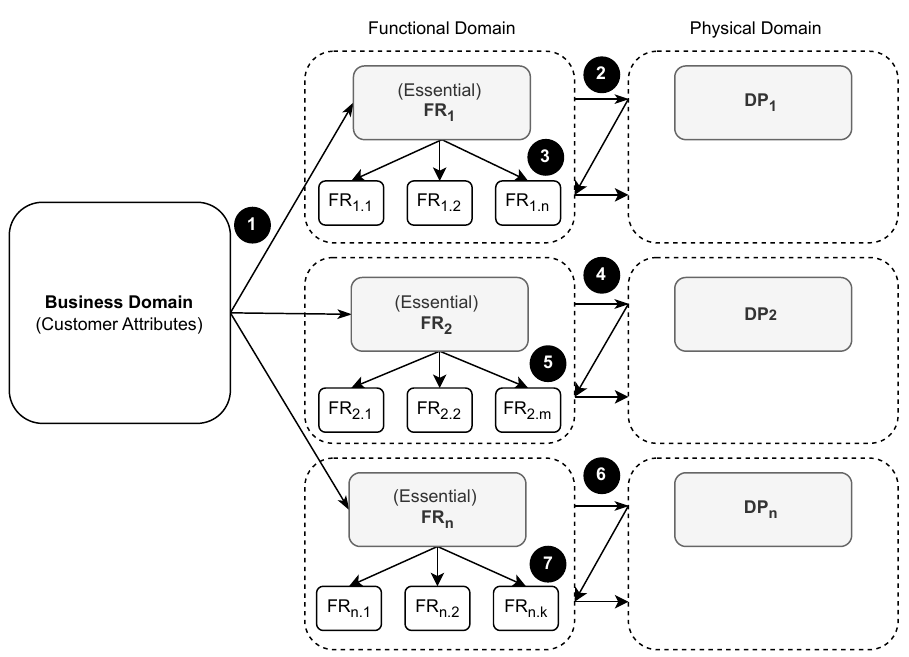}
    \caption{Emergence of essential FRs from $K_W$ and their decomposition into dervied (i.e., lower-level) FRs via zigzagging (knowledge expansion from $K_W \cup K_H$ to $K_D \cup C_{sys}$).}
    \label{fig:essential-frs}
\end{figure}

FRs in axiomatic design are organized hierarchically as a result of ``zigzagging''~\cite[pp.~36-38]{Suh-1990}~\cite[pp.~21]{Suh-2001}~\cite[pp.~27]{Suh-2005}. In this hierarchical arrangement (i.e., decomposition tree), the essential FRs appear at the very top (see Fig.~\ref{fig:essential-frs}). 

These FRs differ from all other FRs in that they arise directly from customer needs and input constraints in a solution-neutral environment~\cite[pp.~15]{Suh-2001}~\cite[pp.~21]{Suh-2005}. This means they establish the functional intent before any commitment to a physical embodiment. This asymmetry explains why preserving independence at the first level is critical—\emph{any coupling introduced here propagates irreversibly} through the decomposition, increasing information content and system complexity in the sense defined by axiomatic design and Suh's complexity theory~\cite{Suh-2001,Suh-2005}.

Essential FRs maintain an \emph{ontological priority} and \emph{causal directionality} in design decomposition, meaning that they define the admissible solution space and constrain the selection of DPs, while remaining invariant to any particular design realization; consequently, changes at this level propagate downstream through all subsequent decomposition layers, whereas modifications in DPs or lower-level FRs cannot retroactively alter them. Essential FRs belong to problem formulation (i.e., the translation of customer needs into a minimum set of independent FRs) and are primarily derived from $K_W$.  

Derived FRs, however, come into existence through \emph{zigzagging} and therefore depend on having (and using) $K_H$ and $C_{sys}$ as well—thus, they are partly shaped by the solution space, not only by needs. In other words, while essential FRs are formulated in a ``solution-neutral'' environment, derived FRs are and will be ``solution-contingent'', meaning that their formulation will depend on the chosen solution (i.e., DPs).

\subsection{Challenges in Formulating Essential FRs}
\label{pitfalls}
As discussed earlier, essential FRs must remain solution-neutral and strictly belong to the functional domain. Preserving their ontological status means that these FRs should describe what the system must achieve, not how it will be realized. Violating this principle introduces premature design decisions and undermines the axiomatic design process from the outset. 

It is important to recognize that while different designers may generate different design solutions for the same set of FRs~\cite[pp.~39]{Suh-2001}~\cite[pp.~56]{Suh-2021-D}, this variability does not extend to the essential FRs themselves. This aligns with the earlier argument that essential FRs capture the essential characteristics of the system of interest.

However, formulating essential FRs involves several challenges. In this section, we summarize the most prominent ones.

\subsection{Identification of Essential FRs}
Although prototyping (when feasible) can help identify essential FRs correctly~\cite[pp.~32]{Suh-1990}, the core idea of hunting for essential FRs is simple: an essential FR is a \emph{non-substitutable} system function—removing it or replacing it with an alternative can change the system’s identity. For example, in the case of a road vehicle, typical essential FRs are ``go forward,'' ``go backward,'' ``stop,'' and ``turn''~\cite[pp.~36--37]{Suh-1990};~\cite[pp.~21]{Suh-2001}; omitting any of them yields an artifact that is no longer a \emph{road vehicle}. Consequently, essential FRs capture the \emph{essential attributes} of customer needs, and \emph{zigzagging} can produce a correct decomposition only when these FRs are identified correctly.

\emph{Theorem S3} reinforces this point by stating that if the first-level FRs are formulated incorrectly, the error cannot be rectified through lower-level design decisions~\cite[pp.~48]{Suh-2005}.

\subsection{Solution-neutrality of Essential FRs}
\label{solution_neutrality}
Although Suh distinguishes between formulating FRs for a new design and improving an existing one (e.g., using the House of Quality to support problem formulation for an existing design)~\cite[pp.~30--35]{Suh-1990}, he generally emphasizes solution-neutral FR formulation at all levels, from the first-level FRs to lower-level FRs~\cite[pp.~15]{Suh-2001}.

However, as stated earlier, maintaining a completely solution-neutral environment during problem formulation is feasible only for the essential FRs. Once zigzagging begins, derived FRs (i.e., derived FRS) can no longer be fully ``solution-neutral'' because they are conditioned by earlier decisions in the physical domain (e.g., by the system constraints those decisions introduce). Therefore, solution-neutral FR definitions are realistically attainable only for the essential FRs, not for derived FRs. The only exception is when the design remains effectively \emph{uncoupled} across levels or when the chosen set of higher level DPs does not constrain or alter the subsequent FRs.

\subsection{Acceptance of Essential FRs}
Constraints provide bounds on the acceptable design solutions~\cite[pp.~39]{Suh-1990}~\cite[pp.~21]{Suh-2001}~\cite[pp.~22]{Suh-2005}. There are two different kinds of constraints: \emph{input} constraints and \emph{system} constraints. 

Input constraints are defined in the customer domain and will be part of the essential FRs. Therefore, it is crucial that essential FRs explicitly absorb input constraints during problem formulation, as delaying their capture to a later stage could result in poor problem formulation. 

System constraints are specific to a given design; they result from design decisions made (i.e., chosen DPs) and are reflected in the derived FRs. 

As noted earlier, the problem formulation at the essential FRs is complete once the input constraints and allowable design ranges have been captured~\cite[pp.~95-96]{Suh-2001}. Let us illustrate this process with an example:

\emph{A customer needs a system that continuously reports the tank level so that the current level is always known. The system must update quickly enough (between 10 ms and 50 ms) to support monitoring and control without delay.}

Below, we present a candidate FR from the problem-formulation step, explicitly capturing the input constraints and the design range:

\noindent Here:  
\begin{itemize}
  \item The ``what'' refers to measuring the tank level.
  \item The input ``constraint'' is continuous execution.
  \item The ``design range'' is \SIrange{10}{50}{\milli\second}.
\end{itemize}

The essential FR could be formulated as: 
\[
FR_{1}: \text{Measure the tank level continuously within } \SIrange{10}{50}{\milli\second}.
\]

Next-level FRs are formulated based on the selected design parameters, and they may incorporate system constraints. In principle, all higher level design decisions will result in the generation of system constraints~\cite[pp.~39]{Suh-1990}~\cite[pp.~21]{Suh-2001}~\cite[pp.~25]{Suh-2005}. Note that optimizing input constraints before problem formulation will be a sound design strategy because it could impact both the quantity and quality of the essential FRs, resulting in overall design optimization.

\subsection{Impact of Creativity on Essential FRs}
A phone’s two primary essential FRs are (1) ``receive a call'' and (2) ``make a call''. If a device does not satisfy both FRs, it cannot be categorized as a phone. Advancements in transmission mediums—most notably the adoption of electromagnetic waves—revolutionized the phone’s attributes by adding another essential FR, (3) ``mobility,'' thereby transforming the phone into a mobile phone.

Creativity can, in some cases, yield truly new essential FRs. When that happens, it can reshape the problem so radically that the resulting design effectively belongs to a different set of business needs. As another example, consider essential FRs for a vehicle before and after the invention of the wheel. Moving ``backward'' became a valued customer attribute only once the wheel made such motion feasible and meaningful.

A particularly illuminating case comes from medical imaging: computed tomography, magnetic resonance imaging, and positron emission tomography share very similar and almost unique essential FRs while differing in constraints and derived FRs.

In most cases, the original needs and/or essential FRs remain a subset of an expanded set of needs and/or essential FRs; in rarer cases, the innovation effectively reformulates business needs and/or essential FRs radically.

\section{Abstract Thinking and What LLMs Can Offer}
\label{abstract_thinking}
Now consider a designer reformulating essential FRs of a phone to design a mobile phone, i.e., formulating ``receive a call'' and ``make a call'' as ``receive a call over electromagnetic waves'' and ``make a call over electromagnetic waves.''. Such statements collapse a contingent design choice (the transmission medium) into the functional domain, prematurely constraining the solution space, violating the independence of FRs by capturing more than one FR in a single FR statement and obscuring the more general essential FRs that define the phone and the mobile phone category. This kind of poor problem formulation can easily be replicated in LLMs. Because abductive reasoning and abstract thinking are complex judgments that are fundamentally beyond the capabilities of current LLMs. 

\subsection{The Critical Role of Abstract Thinkin}
This study argues that during problem formulation in terms of the essential FRs, \emph{abstract thinking} plays a central role because abstract information at the level of problem formulation is more valuable than concrete details. Abstract thinking is a cognitive process that focuses on the superordinate and general characteristics of an event. This form of thinking can enhance creativity because it enables individuals to generate novel and even unprecedented solutions~\cite[pp.~267]{Qin-Wang-2023} and frequently serves as the primary driver of scientific discovery\footnote{The role of abstract thinking in scientific discovery has been examined by many scholars, and our argument draws on landmark studies by Ohlsson and his colleagues~\cite{Regan-Ohlsson-1991,Ohlsson-Lehtinen-1997,Ohlsson-Regan-2001}.}.

Problem formulation in terms of essential FRs will benefit more from abstract information (i.e., conceptual world) rather than dealing with details that may become irrelevant once the design has progressed further.\footnote{Suh maintains that ``a good designer must be able to operate in the \emph{conceptual} world of the functional as well as the physical domain''~\cite[pp.~28]{Suh-1990}. Recent developments in cognitive science support our claim~\cite{Kapoor-Kaufman-2023}.} 

In fact, jumping into design details too early can introduce what Suh calls \emph{imaginary complexity} in his \emph{Complexity Theory}\footnote{Imaginary complexity is defined as uncertainty that is not real uncertainty but arises because of the designer's lack of knowledge and understanding of the specific design itself~\cite[pp.~65]{Suh-2005}.}.

However, the key question is how far the designer should pursue abstract thinking in the functional domain. Advancing abstract thinking in the functional domain too far could hinder the decomposition of business needs into independent essential FRs and instead produce an overly coarse single FR that merely restates the business narrative or mixes the ``what'' with the ``how'' in terms of the business domain vs functional domain.

Remember, the goal of abstract thinking in axiomatic design is not to rephrase customer attributes (i.e., customer needs) but to decompose them into independent functional requirements at an appropriate level while deferring detailed considerations to later stages. The only acceptable level at this stage is the level that results in capturing essential FRs.

To illustrate, consider a customer need for a flying object. The objective is not to restate this need in the functional domain but to decompose it into a set of essential FRs such as ``lift,'' ``thrust,'' and ``maneuver.'' that can explain the ``how'' in terms of the ``what'' in the business domain. Note that one may ask why ``landing'' is not included at this level of abstraction.

The reason is that ``landing'' emerges from the decomposition of ``lift'' into lower-level FRs that regulate lift—either to balance weight and sustain flight or to intentionally reduce lift under controlled conditions for descent and landing. Its formulation therefore depends on the chosen means, representing the ``how'' for realizing the ``what'' defined by ``lift.'' i.e., ``landing'' is addressed through the decomposition of the lifting requirement rather than being introduced as a separate essential FR. Then, depending on the design parameters chosen to satisfy the ``lift'', the next level FRs might be defined to include \emph{landing wheels} or \emph{landing skids}.

This is the degree of abstract thinking expected during problem formulation: capturing the essential functional structure, much like identifying fundamental properties in physics and chemistry (e.g., what defines a fluid—the intrinsic characteristics that make a substance behave as a fluid, independent of composition). Suh describes this step as \textbf{the most difficult part}~\cite[pp.~30]{Suh-1990} of the design process.

\section{What LLMs Can and Cannot}
\label{advancement}

The Abstract Reasoning Corpus (ARC) challenge represents a significant benchmark for developing Artificial Intelligence (AI) systems capable of human-like reasoning. Even the most advanced Large Language Models (LLMs) fall short of human performance~\cite{Lee-Kim-2025}. Although there are works that have attempted to improve this situation~\cite{Pateria-Quek-2022}, at the time of writing, it remains unlikely that state-of-the-art LLMs can correctly identify these essential FRs because they simply cannot reach the level of abstract thinking presented earlier in this study.

On the other hand, this study argues that problem formulation in terms of the essential FRs is an \emph{abductive}/\emph{inductive} process in which the designer uses inference to the best explanation (i.e., abduction) or relies on past experiences (i.e., induction) to propose the most plausible essential FRs, given the observed information (customer needs).

Although LLMs perform well in \emph{induction} and are increasingly capable of \emph{deductive} reasoning, they still struggle with \emph{abduction}~\cite{Pateria-Quek-2022,Lee-Kim-2025,Tom-2026}. This helps explain why GenAI chatbots can assist designers in identifying design parameters and even formulating derived FRs; yet, they often fall short when it comes to deriving the essential FRs.

\section{Conclusion}
\label{Conclusion}
This paper investigates problem formulation in axiomatic design, focusing on how customer needs and input constraints can be converted into the smallest set of independent functional requirements (FRs) called essential FRs. It argues that, for a fixed set of needs and constraints, different designers should arrive at the same essential FRs.

Essential functional requirements (FRs) constitute a minimal, non-substitutable, and solution-neutral set of requirements that emerge from $K_W$ during problem formulation and define the boundary of the design in the functional domain. They possess ontological priority in the design process: they determine the admissible set of design parameters (DPs) at the topmost level while remaining invariant across alternative design realizations as long as customer needs are unchanged. In contrast, derived FRs arise through iterative zigzagging under the influence of $K_D$, $K_H$, and $C_{sys}$, making them inherently solution-contingent and epistemological. This ontological asymmetry implies a unidirectional causal structure in design decomposition, where changes in essential FRs propagate through all subsequent design layers, whereas changes in DPs cannot retroactively alter essential FRs. Consequently, preserving independence at the level of essential FRs is critical, as any coupling introduced at this stage propagates irreversibly, increasing information content and system complexity in the sense defined by axiomatic design.

Abstract thinking is fundamental to the formulation of essential FRs because it enables the designer to extract underlying functional intent from customer needs while avoiding premature commitment to specific solutions. Future work should further examine how essential FRs influence modular design decisions and decomposition optimization.

\bibliographystyle{splncs04}
\bibliography{references}

\end{document}